\def\year{2021}\relax
\documentclass[letterpaper, final]{article} 
\usepackage{aaai21}  
\usepackage{times}  
\usepackage{helvet} 
\usepackage{courier}  
\usepackage[hyphens]{url}  
\usepackage{graphicx} 
\usepackage{graphicx}  
\usepackage{natbib}  
\usepackage{caption} 
\usepackage{todonotes}
\usepackage{multicol}
\usepackage{multirow}
\usepackage{subfig}
\usepackage{mathtools}
\usepackage{fixbib}
\usepackage{bm}
\usepackage{amsmath}
\usepackage{amssymb}
\usepackage{booktabs}
\usepackage{threeparttable}
\usepackage{array}

\newcommand{\specialcell}[2][l]{%
  \begin{tabular}[#1]{@{}c@{}}#2\end{tabular}}

\title{Revisiting Mahalanobis Distance for Transformer-Based Out-of-Domain Detection}
\author{
Alexander Podolskiy\textsuperscript{\rm 1}, Dmitry Lipin\textsuperscript{\rm 1}, Andrey Bout\textsuperscript{\rm 1},\\
Ekaterina Artemova\textsuperscript{\rm 1, 2},  Irina Piontkovskaya\textsuperscript{\rm 1}\\
}

\affiliations{
\textsuperscript{\rm 1} Huawei Noah's Ark Lab, Moscow, Russia
\textsuperscript{\rm 2} HSE University, Moscow, Russia
\text{\{podolskiy.alexander, dmitry.lipin, bout.andrey, artemova.ekaterina, piontkovskaya.irina\}@huawei.com}
}

\begin{document}

\maketitle

\begin{abstract}
Real-life applications, heavily relying on machine learning, such as dialog systems, demand out-of-domain detection methods. Intent classification models should be equipped with a mechanism to distinguish seen intents from unseen ones so that the dialog agent is capable of rejecting the latter and avoiding undesired behavior. However, despite increasing attention paid to the task, the best practices for out-of-domain intent detection have not yet been fully established. 

This paper conducts a thorough comparison of out-of-domain intent detection methods. We prioritize the methods, not requiring access to out-of-domain data during training, gathering of which is extremely time- and labor-consuming due to lexical and stylistic variation of user utterances. We evaluate multiple contextual encoders and methods, proven to be efficient, on three standard datasets for intent classification, expanded with out-of-domain utterances. Our main findings show that fine-tuning Transformer-based encoders on in-domain data leads to superior results. Mahalanobis distance, together with utterance representations, derived from Transformer-based encoders, outperforms other methods by a wide margin and establishes new state-of-the-art results for all datasets. 

The broader analysis shows that the reason for success lies in the fact that the fine-tuned Transformer is capable of constructing homogeneous representations of in-domain utterances, revealing geometrical disparity to out of domain utterances. In turn, the Mahalanobis distance captures this disparity easily.

\end{abstract}

\section{Introduction}

The usability of dialog systems depends crucially on the capability of dialog agents to recognize user intents. Recently deep classifiers have been widely used to recognize user intents, leveraging efficient pre-training, and large amounts of labeled data. However, the scope of annotated corpora is inherently limited, leading to unsatisfactory results when presented with unseen intents. Rather than trying to match user utterances to a limited number of intent classes, dialog agents may be equipped with an auxiliary mechanism to distinguish between seen and unseen intents, i.e., to identify out-of-domain (OOD) utterances. The OOD detection mechanism must handle unseen intents to prevent the erroneous actions of dialog agents.  

Multiple recent papers emphasize the increasing importance of OOD utterances detection caused by the spreading integration of classification models to real-life applications and dialog systems. Simultaneously, in the overwhelming majority of papers, the task is approached in an unsupervised way, see \cite{ll_ratio_nlp_facebook,larson2019evaluation,zheng2020out}.  To this end, the primary approach relies on a decision rule, which is defined to score each utterance.  The scores are further used to reject OOD utterances or to subject to further processing in-domain (ID) ones.  An intuitive yet efficient decision rule determines a threshold for softmax output probabilities, measuring the classifier's confidence. The less confident the classifier is, the higher are the chances to reject the utterance. Other decision rules rely on distance-based approaches to check whether an utterance falls out of ID space. 

With Transformer-based contextual encoders becoming core to almost, if not all, NLP methods, undoubtedly, their performance for intent classification is well-studied. However, the performance of Transformers in the OOD detection task so far has been little explored. \citet{Hendrycks2020PretrainedTI} provide evidence that Transformers generalize well to unseen domains in sentiment classification and sentence pair modeling tasks, suggesting that Transformers will perform better than previous models for the task of OOD utterance detection, too. This paper fills this gap in the evaluation of Transformers. 

The key idea of this paper is to conduct a comprehensive comparison of the performance of different contextual encoders in multiple settings. We adopt three dialog datasets designed for the task of OOD intent detection along with the current best practices and state-of-the-art methods for the task. Although Transformers primarily outperform other contextual encoders by a wide margin, they serve as especially useful embedders for the distance-based methods of OOD detection. When fine-tuned on ID data, Transformers form dense clusters of ID utterances, which are easy to locate with Mahalanobis distance. 

To summarize, the key contributions of the paper are as follows:

\begin{enumerate}
    \item We evaluate multiple contextual encoders and best practices for OOD detection on three common datasets for intent classification, expanded with out-of-domain utterances;
    \item  We show that not only fine-tuning Transformers on ID data   consistently improves OOD detection, but also that when combined with Mahalanobis distance, it established new state-of-the-art results;
    \item We discover that the fine-tuned Transformer is capable of constructing homogeneous representations of ID utterances, revealing geometrical disparity to OOD ones, captured easily in turn by the Mahalanobis distance.

\end{enumerate}

\section{Related Work}

Methods for OOD detection can be roughly grouped based on whether they have access to OOD data and whether they utilize ID labels. 

{\bf Classification methods} require access to OOD data for supervision.  \citet{larson2019evaluation} use two supervised settings: 1) binary classification, so that all ID classes are classified against OOD one, and 2) training an additional class for OOD inputs.
\cite{kamath2020selective} train an additional model,  a {\bf calibrator}, which identifies inputs on which the classifier errs, and rejects those inputs, for which an error is likely. \cite{hendrycks2018deep_OE}  utilize OOD data for outlier detection by training models to increase entropy on OOD examples.

However, in real-life applications gathering and maintaining OOD data is complicated by the lexical and stylistic variation of user utterances \cite{rostd_indomain}. For this reason, methods without OOD supervision gain more attention.

{\bf Outputs of the classifier},
trained with the supervision of ID classes, can be exploited as a score for OOD inputs. Maximum softmax probability  \cite{msp}  is recognized as a strong baseline, when used with deep classifiers, improved further by introducing temperature scaling \cite{odin}. KL-divergence captures the changes in prediction distributions learned for an ID class by the classifier and detects the arbitrary guesses made for OOD inputs \cite{yilmaz2020kloos}.

{\bf Generative methods} use a natural ability of language models and other generative models to estimate the likelihood of the inputs \cite{nalisnick2018do,ll_ratio_google}. \cite{zheng2020out} utilize ID inputs and unlabeled data to generate pseudo-OOD utterances with a generative adversarial network, improving  OOD detection on a dialog dataset.

{\bf Distance-based methods} treat distance estimation as an OOD score: the further an input is from ID inputs, the higher are the chances that it is OOD \cite{Mandelbaum2017DistancebasedCS,Gu2019StatisticalAO,Lee2018ASU}.

Other research direction include  {\bf Bayesian estimation for uncertanity} derived from learned distributions over network weights  \cite{NIPS2018_7936, blundell2015weight}, processing of {\bf lexical features} \cite{ghosal2018investigating} and training prototypical networks  to define class prototypes for each ID class \cite{tan2019out}.

\section{Background}

Let $\mathcal{D}_{ID} = \{(x_1, y_1),\dots, (x_n, y_n)\}$ be a dataset, where $x_i$ is an input utterance and $y_i \in \Upsilon$ is its class label. Than $\Upsilon$ is the set of seen, in-domain classes, and the total number of classes is $\vert\Upsilon\vert = N$. Assume that ID utterances are drawn from the distribution $P_{ID}$ and that there exists an OOD distribution, $P_{OOD}$, which differs from $P_{ID}$. Finally, suppose that a scoring function  maps an utterance $x$ into a real number. The OOD detector then accepts the ID utterances and rejects the OOD utterances according to the decision rule in Eq.~\ref{eq:OOD_descion_rule}.

\begin{equation}
        R(x)= 
\begin{dcases}
    \textbf{reject} ,& \text{if } d(x)\geq \theta\\
    \textbf{accept},              & \text{otherwise}
\end{dcases}
\label{eq:OOD_descion_rule}
\end{equation}
where $\theta$ is a threshold, $d$ can be either independent from $y$, otherwise model joint $d(x,y)$ or conditional $d(x | y)$ dependence.  Ideally, we want $d(x) < d(\hat{x})$ for all $x\sim P_{IN}$, $\hat{x}\sim P_{OOD}$.

\subsection{Methods}

We adopt several methods that do not rely on access to OOD data and are shown to be effective for OOD detection in vision and natural language tasks. We exploit Maximum Softmax Probability (MSP) as a strong baseline \cite{msp}, Likelihood ratio \cite{ll_ratio_nlp_facebook} as the current state-of-the-art method for dialog data. We use Mahalanobis distance \cite{de2000mahalanobis}, an advanced distance-based method, computed in multiple ways. It is the most straightforward to compute the Mahalanobis distance to the closest ID classes, assuming that the ID labels are provided. If not, marginal Mahalanobis distance allows computing the distance to the ID data centroid.

\textbf{Maximum Softmax Probability (MSP) } requires a pre-trained classifier $f$ with a softmax output layer \cite{msp}. Let $p_y(x)$ denote the probability, assigned by $f$, to the utterance  $x$ to belong to class $y$. The less classifier is confident with its prediction, the higher is the OOD score:
\begin{equation}
    d(x) = 1 - \max\limits_{y\in \Upsilon} p_y(x).
\label{eq:msp}
\end{equation}

To prevent the classifier from  becoming too confident in its prediction,  \cite{odin} introduce softmax temperature scaling at the test time: 
$p_y(x) = {e^{\frac{z_y}{\tau}}}/{\sum\limits_{y\in \Upsilon} e^{\frac{z_y}{\tau}}}$
, where $z_y$  denotes the logit for label $y$ while $\tau$ denotes the softmax temperature.

\textbf{Likelihood Ratio (LLR)} exploits two language models \cite{ll_ratio_nlp_facebook}. One of them, $L(x)$, is trained on source data and aims to capture ID utterances' semantics. The second language model, $L_{bg}$, addressed as a background model, is trained on corrupted with some noise source data and aimed at learning the background statistics. The final score is computed as follows in Eg.~\ref{eq:LLR}.

\begin{equation}\label{eq:LLR}
    d(x) = -\log\frac{L(x)}{L_{bg}(x)},
\end{equation}

\textbf{Mahalanobis distance} is a way to determine the closeness of an utterance to a set of utterances belonging to the class $c$. 
Following \cite{Lee2018ASU}, we define Mahalanobis distance, serving as OOD score, as:
\begin{equation}\label{eq:Maha}
\begin{gathered}
    d(x) = \min\limits_{c\in\Upsilon}(\psi(x) - \mu_c)^T\Sigma^{-1}(\psi(x) - \mu_c),
\end{gathered}
\end{equation}
where $\psi(x)$ is a vector representation of the utterance $x$, $\mu_c$ is the centroid for a class $c$ and $\Sigma$ is the co-variance matrix. The estimations of $\mu_c$ and $\Sigma$ are defined by

\begin{equation*}
    \begin{gathered}
            \mu_c = \frac{1}{N_c}\sum\limits_{x\in\mathcal{D}^c_{in}}\psi(x),\\
        \Sigma = \frac{1}{N}\sum\limits_{c\in\Upsilon}\sum\limits_{x \in\mathcal{D}^c_{in}}(\psi(x) - \mu_c)(\psi(x) - \mu_c)^T,
    \end{gathered}
\end{equation*}
where $\mathcal{D}^c_{IN} = \{x \vert (x, y)\in\mathcal{D}_{in},\, y = c\}$, $N$ is the total number of utterances, and $N_c$ is the number of utterances belonging to class $c$.

\section{Datasets}

To the best of our knowledge, we are the first to evaluate OOD detection with three NLU datasets, consisting of both ID and OOD utterances. 

{\bf CLINC150} is an intent classification dataset, modeling a real-life situation. Some utterances fall out of domains, covered by train data \cite{clinc150}. The total number of ID classes in CLINC150 is equal to 150. The OOD utterances relate to actions not supported by existing ID intents.  

{\bf ROSTD} extends the English part of multilingual dialog dataset with OOD utterances \cite{rostd_indomain,ll_ratio_nlp_facebook}. The hierarchical label structure of ROSTD allows us to experiment with both a larger number of classes (12) or ``coarsened'' classes (3). Following \cite{ll_ratio_nlp_facebook}, we experiment with both variants and refer to them as ROSTD and ROSTD-COARSE.  The OOD part consists mainly of subjective, under-specified, or over-emotional utterances that do not fall into ID classes.

{\bf SNIPS} has no explicit ID/OOD split. The total number of intents is 7. Following \cite{intent_detection_margin_loss} setup,  we randomly split all labels into ID and OOD parts. The ID part covers about $75\%$ of the whole dataset. We average the results of all splits. 

Table~\ref{tab: data_stats} presents with dataset statistics.

\begin{table}[!ht]
\centering
\begin{tabular}{cccc}
\toprule
                 & CLINC150 & ROSTD & SNIPS \\
\midrule
Number train IND & 15K    & 30K     & 13K   \\                                  
Number val IND   & 3K     & 4K     & 0.7K  \\                           
Number val OOD   & 0.1K   &  1.5K    & --  \\                                   
Number test IND  & 4.5K   & 8.6K     & 0.7K \\                    

Number test OOD  & 1K     & 3K     & --  \\       
\bottomrule

\end{tabular}
\caption{Dataset statistics}
\label{tab: data_stats}
\end{table}

\section{Embeddings and encoders}

We evaluate three representation models, ranging from bag-of-words, static pre-trained word embeddings up to contextualized encoders. 

{\bf Bag-of-words.} We use the bag-of-words model \cite{haris_bow}, which shows stable performance due to its low variance.

{\bf Static word embeddings.} We use GloVe \cite{pennington2014glove} as inputs to a convolutional neural network (CNN) and long short-term memory (LSTM), trained further with the supervision of ID data. The CNN architecture follows one used in \cite{zheng2020out}.  We use GloVe vectors as inputs to language models needed for LLR. LSTM is used as an underlying model of LLR, trained on ID data with language modeling objective. We train the background model on the ID data with added uniform noise. We find that 0.5 noise probability performs the best. 

{\bf Pre-trained Transformers.} We utilize multiple BERT-based models \cite{devlin2019bert}, which are pre-trained Transformers, trained with a self-supervised masked language modeling objective. Additionally to BERT-base and BERT-large, we use RoBERTa-base and RoBERTa-large models \cite{liu2019roberta}. We use distilled versions of both BERT and RoBERTa, DistillBERT and DistillRoBERTa \cite{sanh2019distilbert}.

Each CNN, LSTM, and Transformer model is used as a classifier with the MSP method and as an embedder with Mahalanobis distance. We follow the standard fine-tuning procedure to fine-tune each model for three ID intent classification tasks.  We tune hyper-parameters to maximize performance on the validation set for each of the ID intent classification tasks. We perform our experiments with PyTorch \cite{paszke2019pytorch}, PyTorch Lightning \cite{falcon2019pytorch} and Hugging Face Transformers library \cite{wolf2019huggingface}

\section{Evaluation}

\begin{table*}[ht]
\centering
\begin{threeparttable}
\begin{tabular}{cccccc}
\toprule
Dataset & Model & $\rm{AUROC}\uparrow$ & $\rm{AUPR_{OOD}}\uparrow$ & $\rm{FPR@95_{OOD}}\downarrow$ & $\rm{FPR@95_{ID}}\downarrow$ \\
\midrule
\multirow{9}{*}{CLINC150} & BoW MSP & $91.5\pm0.0$ & $66.7\pm0.2$ & $31.7\pm0.4$ & $43.9\pm0.9$ \\
& LSTM MSP & $90.9\pm0.6$ & $67.8\pm2.1$ & $31.2\pm2.0$ & $50.7\pm3.0$ \\
&   CNN MSP  & $94.1\pm0.6$ & $80.8\pm2.1$ &     $26.4\pm4.0$       &     $24.4\pm2.8$      \\
& CNN Maha & $95.2\pm0.2$ & $76.2\pm1.4$ & $16.4\pm1.1$ & $27.8\pm1.6$ \\
& LLR & $91.4\pm0.3$ & $73.1\pm1.0$ & $37.0\pm1.5$ & $39.9\pm1.5$ \\
& $\rm{BERT_{\rm base}}$ Maha       &   $97.3\pm0.1$    &    $88.6\pm1.0$     &     $10.9\pm0.7$       &     $12.5\pm1.1$      \\
& $\rm{BERT_{\rm base}}$ SNGP\tnote{1} & $96.9\pm1.0$ & $88.0\pm1.0$ & -- & -- \\
& RoBERTa MSP & $97.1\pm0.6$ & $91.2\pm1.3$ & $11.6\pm2.4$ & $12.5\pm2.2$ \\
& RoBERTa Maha & \bm{$98.4\pm0.1$} & \bm{$94.5\pm0.5$} & \bm{$6.8\pm0.8$} & \bm{$7.3\pm1.1$} \\
\midrule
\multirow{7}{*}{ROSTD} & BoW MSP & $94.2\pm0.1$ & $86.7\pm0.1$ & $30.5\pm0.4$ & $25.8\pm0.2$ \\
& LSTM MSP & $73.7\pm8.3$ & $60.6\pm12.1$ & $63.0\pm6.0$ & $57.4\pm13.8$ \\
&  CNN MSP  &  $95.2\pm1.2$     &    $88.2\pm2.8$     &  $22.2\pm6.3$ & $32.5\pm6.0$ \\
& CNN Maha & $98.1\pm0.2$ & $93.3\pm0.7$ & $7.6\pm1.5$ & $7.8\pm1.3$ \\
& LLR & $97.7\pm0.2$ & $95.6\pm0.3$ & $12.3\pm1.7$ & $9.3\pm1.0$ \\
& RoBERTa MSP & $99.3\pm0.2$ & $98.2\pm0.4$ & $2.2\pm0.8$ & $1.8\pm0.9$ \\
& RoBERTa Maha & \bm{$99.8\pm0.1$} & \bm{$99.5\pm0.3$} & \bm{$0.5\pm0.4$} & \bm{$1.0\pm0.5$} \\
\midrule
\multirow{7}{*}{ROSTD-coarse} & BoW MSP & $98.0\pm0.1$ & $96.0\pm0.1$ & $7.8\pm0.7$ & $6.6\pm0.2$ \\
& LSTM MSP & $86.3\pm7.8$ & $80.2\pm10.6$ & $52.7\pm13.5$ & $32.0\pm15.3$ \\
&  CNN MSP  & $97.0\pm0.8$  & $94.7\pm1.3$ & $19.8\pm8.1$ & $10.4\pm2.7$ \\
& CNN Maha & $99.0\pm0.2$ & $97.5\pm0.4$ & $4.5\pm1.1$ & $4.6\pm0.8$ \\
& LLR & $97.7\pm0.2$ & $95.5\pm0.4$ & $12.5\pm1.4$ & $9.1\pm0.9$ \\
& RoBERTa MSP & $99.2\pm0.5$ & $98.8\pm0.5$ & $0.6\pm0.5$ & $1.7\pm0.9$ \\
& RoBERTa Maha & \bm{$99.8\pm0.1$} & \bm{$99.6\pm0.1$} & \bm{$0.2\pm0.1$} & \bm{$0.7\pm0.4$} \\
\midrule
\multirow{7}{*}{SNIPS 75}  & BoW MSP & $92.4\pm2.0$ & $76.9\pm6.8$ & $30.7\pm4.3$ & $41.6\pm6.3$ \\
& LSTM MSP & $81.7\pm10.9$ & $59.6\pm15.3$ & $49.9\pm24.7$ & $59.0\pm15.3$ \\
&   CNN MSP  & $93.7\pm2.3$ & $78.7\pm9.1$ & $24.4\pm9.1$ & $20.2\pm8.8$ \\
& CNN Maha & $87.1\pm9.4$ & $75.4\pm12.6$ & $49.3\pm33.6$ & $37.8\pm18.7$ \\
& LLR & $83.5\pm5.2$ & $61.3\pm12.9$ & $65.1\pm16.0$ & $58.1\pm9.0$ \\
& RoBERTa MSP & $95.3\pm2.8$ & $85.7\pm5.6$ & $25.5\pm22.3$ & $18.2\pm9.1$ \\
& RoBERTa Maha & \bm{$97.6\pm1.9$} & \bm{$92.9\pm5.4$} & \bm{$12.3\pm10.3$} & \bm{$11.2\pm10.5$} \\
\bottomrule
\end{tabular}
\caption{Comparison of OOD detection performance. Each result is an average of 10 runs. $\uparrow$ -- greater is better, $\downarrow$ -- lower is better}
\label{tab:all_results}
\begin{tablenotes}
\item[1] Results are taken from \cite{liu2020simple}
\end{tablenotes}
\end{threeparttable}
\end{table*}

The task of OOD detection is a binary classification task, where OOD utterances should be distinguished from ID utterances. In the unsupervised setting, a scoring function is used to assign an OOD score.

$\rm{\mathbf{AUROC}}$, the area under the Receiver Operating Characteristic, can be interpreted as the probability of randomly sampled ID utterance having a lower OOD score than randomly sampled OOD one.

$\rm{\mathbf{{AUPR}_{OOD}}}$, the area under Precision-Recall Curve, requires taking OOD as the positive class. It is more suitable for highly imbalanced data in comparison to $\rm{AUROC}$.

$\rm{\mathbf{FPR@X}} $ corresponds to False Positive Ratio with the decision threshold is set to
$
 \theta = \sup \{\tilde\theta \in \mathbb{R} \mid \rm{TPR}(\tilde\theta) \leq X \}
$
where $\rm{TPR} \in [0,1]$ is a True Positive Rate. This metric also requires selecting one class as positive. Different approaches are used, e.g. \citet*{ll_ratio_nlp_facebook} treat the OOD class as positive one, while \citet*{zheng2020out} choose ID class. We report both metrics:  $\rm{FPR@X}_{ID}$ means that ID class is treated as positive, and $\rm{FPR@X}_{OOD}$ means the same for OOD class.

Two metrics, $\rm{AUROC}$ and $\rm{AUPR}_{OOD}$ are threshold-independent. $\rm{FPR@X} $ requires picking a threshold.

\section{Out-of-Domain Detection}
\label{sec: experiments}
\subsection{Transformers with Mahalanobis distance  are better at OOD detection than other models}

Table~\ref{tab:all_results} presents with the results of experiments. On all datasets, $\rm{RoBERTa}$ equipped with the Mahalanobis distance outperforms baselines, and other methods, including $\rm{RoBERTa}$ with the MSP score. Advantages are even more evident for CLINC150, which is less lexically and syntactically diverse and challenging. 

On the CLINC150 dataset, Mahalanbois distance combined with  Transformer-based embeddings outperforms recently proposed BERT SNGP (Spectral-normalized Neural Gaussian) \cite{liu2020simple}. In order to make a fair comparison, we show the performance of the $\rm{BERT_{base}\ Maha}$.

LSTM with MSP performs at the baseline level and is outperformed with CNN with MSP, followed by the previously established state-of-the-art method, LLR \cite{ll_ratio_nlp_facebook}.
In turn, it does not cope well with CLINC150 and SNIPS and is slightly outperformed by CNN with Mahalanobis distance. LLR might be challenging to apply, as the background model can still learn semantics from the data, even though it is trained on the noisy inputs. There is a high variance in the background model training due to the extensive vocabulary size. 
The Mahalanobis distance and its variants depend primarily on the embeddings learned by a model. All models equipped with MSP were fine-tuned using cross-entropy loss for intent classification.  The Table~\ref{tab:all_results} confirms that such fine-tuning does not always help the models generate informative embeddings. CNN with the Mahalanobis distance shows moderate performance on the ROSTD dataset and its coarse version. However, performance severely drops on more challenging datasets.

\begin{table*}[ht]
\center
\begin{tabular}{>{\centering}p{3.5cm}cccccc}
    \hline
    & \multicolumn{3}{c}{RoBERTa} & \multicolumn{3}{c}{CNN}\\
    \hline
      & CLINC150 & ROSTD & SNIPS & CLINC150 & ROSTD & SNIPS \\
    \specialcell{Pairwise similarity\\ between centroids} & $0\pm0.08$ & $-0.05\pm0.13$ & $-0.16\pm0.12$ & $0.35\pm0.11$ & $0.24\pm0.09$ & $0.15\pm0.02$ \\
    \specialcell{Centroid length} & $19.75\pm0.23$ &  $18.09\pm0.36$ & $18.57\pm0.64$ & $23.07\pm1.38$ & $19.12\pm 1.11$ & $18.29\pm1.08$ \\
    \specialcell{Similarity between ID\\ instances and centroids} & $0.96\pm0.11$ & $0.95\pm0.08$ & $0.98\pm0.04$ & $0.92\pm0.08$ & $0.96\pm0.44$ & $0.99\pm0.05$ \\
    \hline
\end{tabular}
\caption{Descriptive statistics of embedding space. Both spaces are derived from fine-tuned models with ID supervision. We show statistics for only one of the SNIPS splits}
\label{tab:center_stats}
\end{table*}

\subsection{Semantically close ID and OOD classes are often confused}

$\rm{RoBERTa}$ with the Mahalanobis distance score is affected by multiple factors.

Mislabeled instances cause top errors made for CLINC150, e.g. {\it give me the weather forecast for today} is labeled as OOD but is related to the intent $\mathbf{weather}$. Similarly, an OOD utterance {\it how old is Jennifer Anniston?}  is incorrectly assigned with the intent {\bf how old are you?}, used to question about the dialog agent's personality.

Other errors include confusion between semantically related utterances.  For example, CLINC150 contains intent $\mathbf{text}$ that is related only to sending a text message. Utterances related to similar actions, such as {\it read my friend's text message}, are erroneously accepted.

Similar issues appear in SNIPS. The structure of ID-OOD  splits explains the high variance of metrics. If two semantically related or often confused intents get into different sets, the resulting measures drop significantly. For example, it is  challenging  if the intent {\bf SearchScreenEvent} is ID and {\bf SearchCreativeWork} is OOD. The rest of the ID intents do not provide enough supervision to learn the former intent's exact semantics to more clearly separate from the latter. 

On the ROSTD dataset, we observe the same errors caused by the semantic similarity between an OOD utterance and ID intents. Another source of errors is the lexical discrepancy between ID utterances in the train and test sets.

\subsection{Is bigger model better?}

\begin{figure}[t]
  \includegraphics[width=\columnwidth]{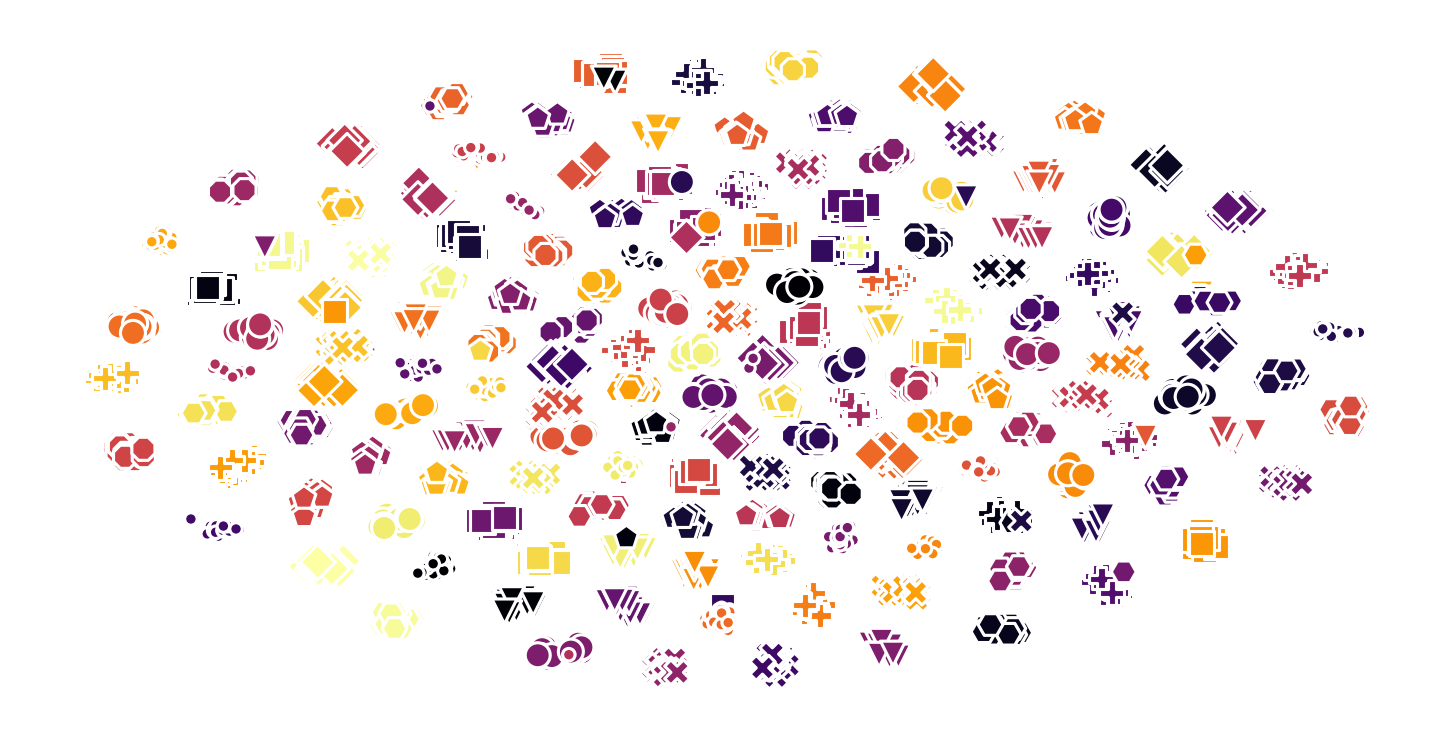}
  \caption{t-SNE visualization of CLINC150 ID classes. Embeddings are derived from fine-tuned RoBERTa for ID classification. ID classes are easily separated}
  \label{fig:tsne}
\end{figure}

\begin{figure}[ht]
    \centering
    \includegraphics[width=8cm]{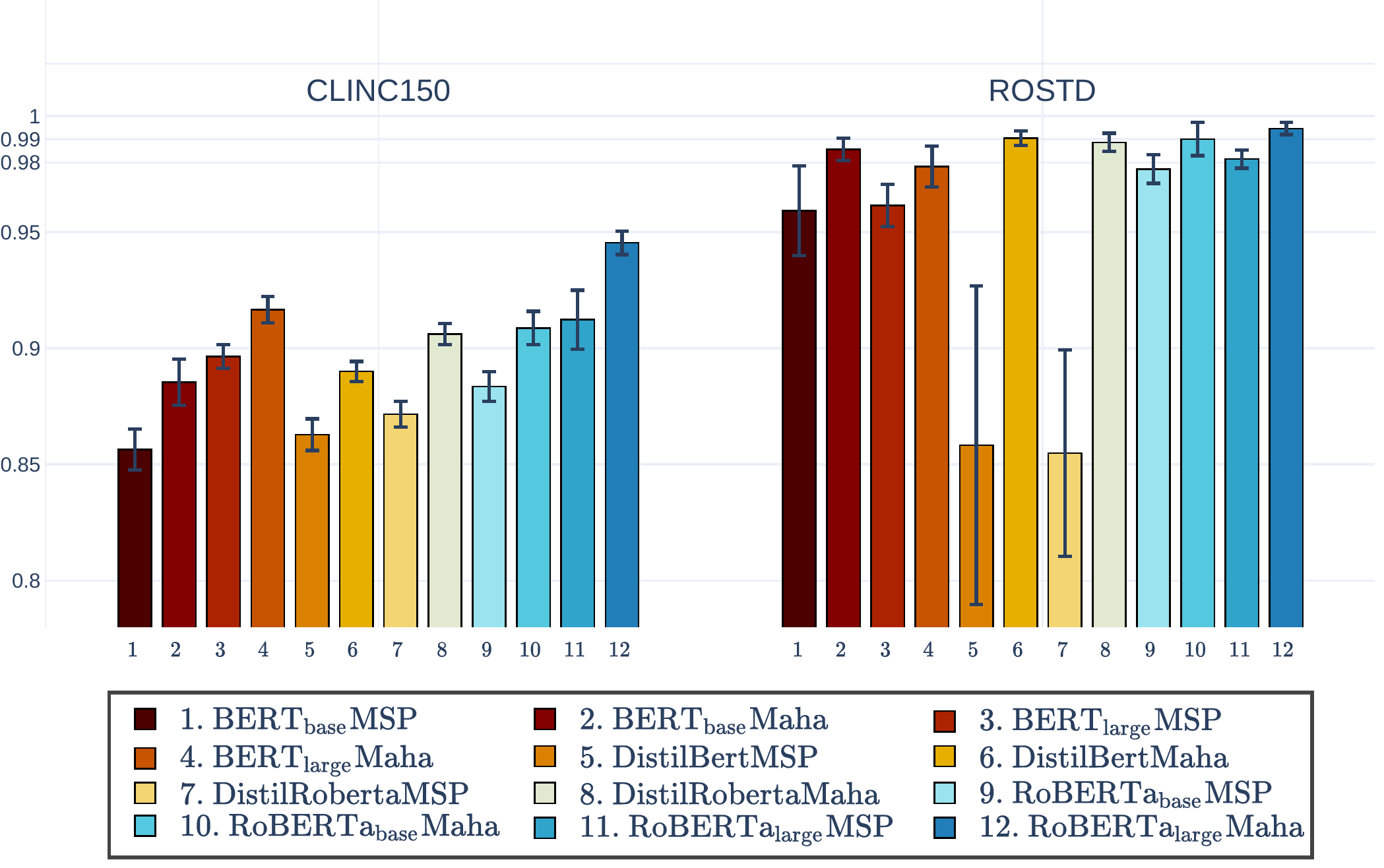}
    \caption{Comparison of models of different sizes on ROSTD and CLINC150. Maha stands for Mahalanobis distance}
    \label{fig:model_size_aupr_ood}
\end{figure}

We utilize base, large, and distilled versions of BERT-based models. Fig.~\ref{fig:model_size_aupr_ood} compares the performance of models with different sizes on two datasets, CLINC150 and ROSTD. On a more diverse dataset, CLINC150, we see that larger models outperform smaller versions. On the ROSTD dataset, this difference is not so prominent but persists. By comparing distilled versions with their respective teachers, we note that the distillation does not affect the Mahalanobis distance, unlike the MSP score. Hence, Mahalanobis distance is more robust to distillation than MSP. 

\subsection{Diverse pre-training data improves OOD detection}

 Fig.~\ref{fig:model_size_aupr_ood} shows that RoBERTa has better OOD detection capabilities than BERT. The core difference between the two models is that RoBERTa was pre-trained on a larger and more diverse dataset than BERT. Thus, we hypothesize that pre-training on larger amounts of data improves model robustness to OOD instances. Recent studies confirm this effect in computer vision \cite{hendrycks_pretraining19,orhan2019robustness} and natural language processing \cite{Hendrycks2020PretrainedTI}.
 
\section{Features of embeddings space}

\begin{figure*}[!ht]
 \centering
    \subfloat[][$\mathrm{RoBERTa}$, no fine-tuning]
    {\includegraphics[width=5.5cm]{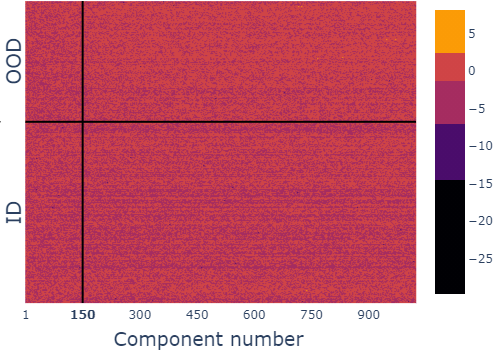}} 
    \hspace{10pt}
    \subfloat[][$\mathrm{RoBERTa}$, fine-tuned ]{\includegraphics[width=5.5cm]{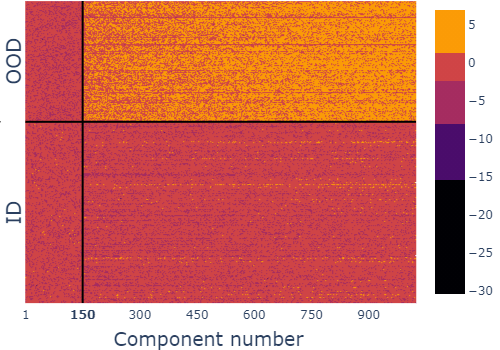}} 
    \hspace{10pt}
    \subfloat[$\mathrm{CNN}$, trained ]{\includegraphics[width=5.5cm]{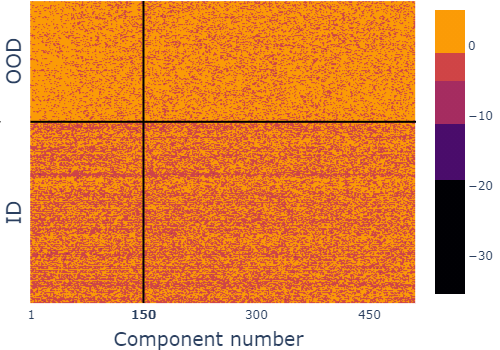}}

    \caption{These heatmaps represent each utterance from the CLINC150 test set as the vector of Mahalanobis distance terms, computed according to Eq.~\ref{eq:Maha_via_PCA} and sorted in the decreasing order of explained variance. Each row stands for an utterance. The horizontal solid line separates the OOD utterances (above the line) from the ID ones (below the line). The vertical solid line splits each heatmap into two parts: to the left are components numbered lower than 150, to the right are components numbered above 150. 150 is the number of classes in the CLINC150 dataset. Only fine-tuned RoBERTa-based vectors clearly distinguish ID and OOD utterances. The difference between ID and OOD is less evident in (c) and almost indistinguishable in (a). However, in (b), the values of the components, starting from the 150$^{th}$ one (in yellow), are lower than those of ID ones (in red). }

    \label{fig:maha_components}
\end{figure*}

Transformer models, fine-tuned on ID data, coupled with Mahalanobis distance, show excellent performance for OOD detection. A possible explanation could be the way the space of Transformer-based embeddings is settled. Further, we compare geometrical features of two different embedding spaces, derived from RoBERTa model and CNN for comparison (see Table~\ref{tab:center_stats}). The differences between the spaces are even sharper if the number of ID classes is high.

{\bf ID class centroids are mutually orthogonal.}  The pairwise cosine similarity between centroids approaches zero as its mean value and standard deviation are close to zero. This reveals that the centroids are mutually orthogonal, as all angles are close to $\frac{\pi}2$. This phenomenon is present for the space of Transformer-based embeddings and does not hold for CNN-based embeddings.

{\bf ID class centroids lay on a sphere.} The length of Transformer-based centroids  does not vary much, as the deviation from the sphere is less than 2\% of its radius. On the other hand, CNN embeddings deviate more significantly.

{\bf ID classes form clusters around centroids.} The deviation of ID instances from the centroids according to cosine similarity is small both for CNN and RoBERTa embeddings. ID data is well clustered, and the classes are well separated from each other, as depicted in Fig.~\ref{fig:tsne}.


{\bf ID data can be approximated by low-dimensional subspace in the embedding space}, because ID embeddings are close to class centroids, and the number of classes is significantly lower than the dimension of the embeddings ($N \ll d$). 

For further analysis, we consider several Mahalanobis distance variants. Following \citet*{Kamoi2020WhyIT}, we introduce the equivalent Mahalanobis distance form, based on  Principal Component Analysis of the class-wise centered ID data:


\begin{equation}
\begin{gathered}
d(\psi(x))=\min\limits_{c}\sum\limits_{i=1}^{d}\frac{y_i^2(\psi(x) - \mu_c)}{\lambda_i},
\end{gathered}
\label{eq:Maha_via_PCA}
\end{equation}

where $y_i(\psi(x))$ is the $i$-th component of the PCA transform of $\psi(x)$, $\lambda_i$ are explained variances of the corresponding principal components, $\mu_c$ are class centroids.

\citet*{Kamoi2020WhyIT} introduced two modifications of Eq.~\ref{eq:Maha_via_PCA}, namely, {\bf marginal Mahalanobis distance}, which  ignores class information and uses instead a single mean vector for all ID classes, (see Eq.~\ref{eq:M_Maha}) and {\bf partial Mahalanobis distances}: it is the version of the equations (\ref{eq:Maha_via_PCA}) and (\ref{eq:M_Maha}) with the summation starting from $N$-th component. Eq. \ref{eq:P_Maha} corresponds to the {\bf partial marginal} variant. Marginal Mahalanobis distance aims at using more compact data representation in the form of a single ID centroid, helping to reduce the amount of data needed for OOD detection. Partial variant utilizes  the most important terms only.

\begin{equation}
\begin{gathered}
        d(\psi(x))=\sum\limits_{i=1}^{d}\frac{y_i^2(\psi(x)-\mu)}{\lambda_i}
\end{gathered}
\label{eq:M_Maha}
\end{equation}

\begin{equation}
\begin{gathered}
        d(\psi(x), N) = \sum\limits_{i=N}^{d} \frac{y_i^2(\psi(x)-\mu)}{\lambda_i}, 
\end{gathered}
\label{eq:P_Maha}
\end{equation}
where
$$
\mu  = \frac{1}{N}\sum\limits_{x\in\mathcal{D}_{in}} \psi(x), 
$$
stands for the ID data centroid.

{\bf Mahalanobis distance can efficiently utilize low-dimensional nature of ID data.} 
 Following the properties of PCA \cite{murphy2012machine}, if the data is approximately $N$-dimensional, it is explained by the first $N$ principal components. That means that for ID data, all the terms in the Eq. \ref{eq:M_Maha}, \ref{eq:P_Maha} are little, while OOD data can be detected by important loadings of the terms $\frac{y_i^2}{\lambda_i}$ with $i>N$.
 To check this, we plot the terms of the Eq. \ref{eq:P_Maha} for ID and OOD data, Fig. \ref{fig:maha_components}. Fig. 3 shows that when decomposed with the Mahalanobis distance embeddings of fine-tuned RoBERTa fall into two parts. The last components of OOD embeddings have a higher variance when compared to the first ones. This phenomena is observed neither for ID embeddings nor for RoBERTa without fine-tuning nor for the trained CNN. 

{\bf Comparison of other distances.}

 We compare Mahalanobis distance variants to explore this matter: original, marginal Mahalanobis, and their partial versions. Additionally, we exploit Euclidean distance to complete our evaluation.

\begin{figure}[ht]
\centering

  \begin{subfloat}
    \centering\includegraphics[width=7.5cm]{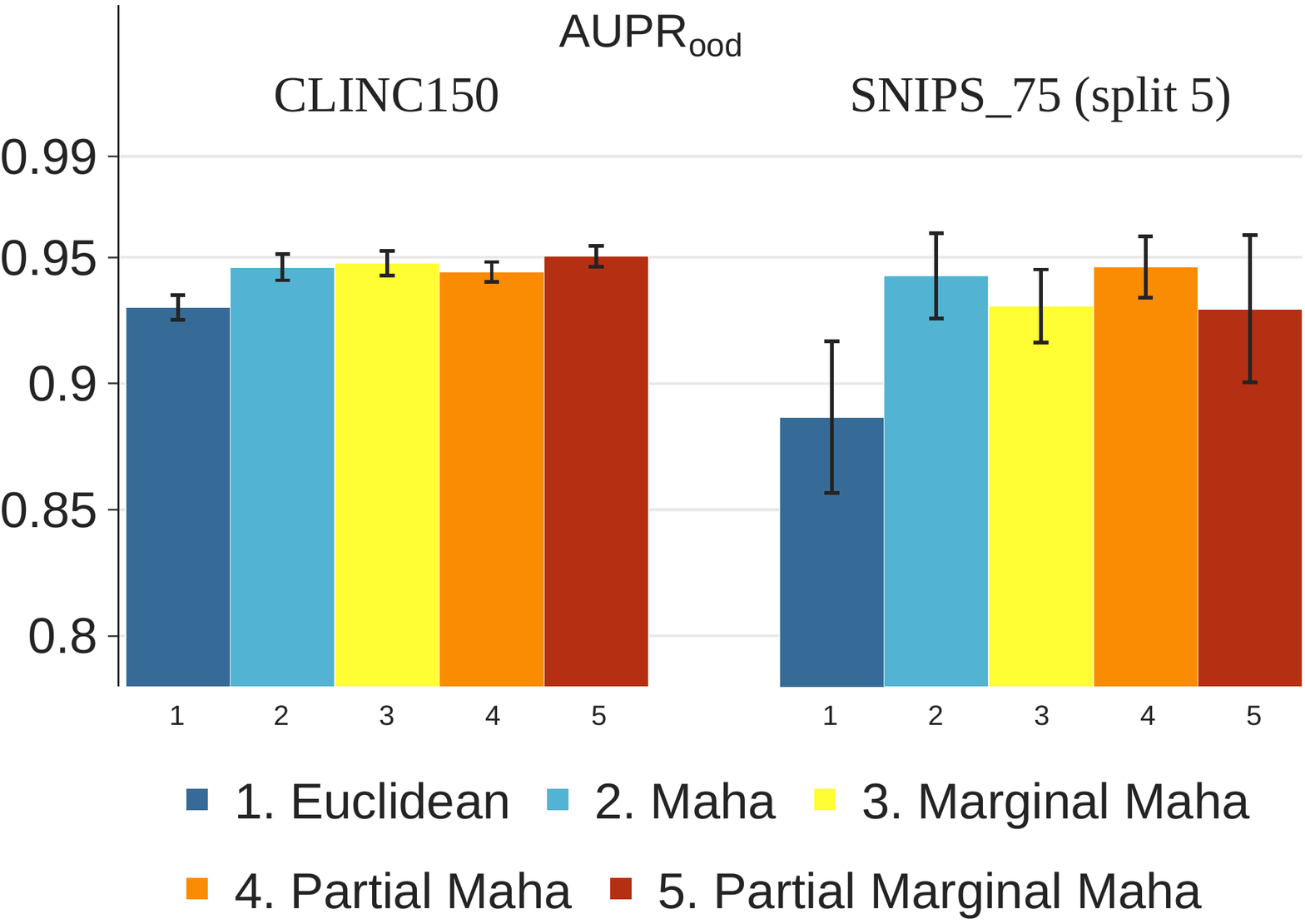}
    \caption*{}
  \end{subfloat}
  \leavevmode\\[-8ex]

  \caption{Comparison of different distances. Mahalanobis distance and its variants outperform Euclidean distance by a wide margin.}
  \label{fig:maha_var_comp}
\end{figure}

\begin{figure}[t]
    \centering
    \includegraphics[width=8cm, keepaspectratio]{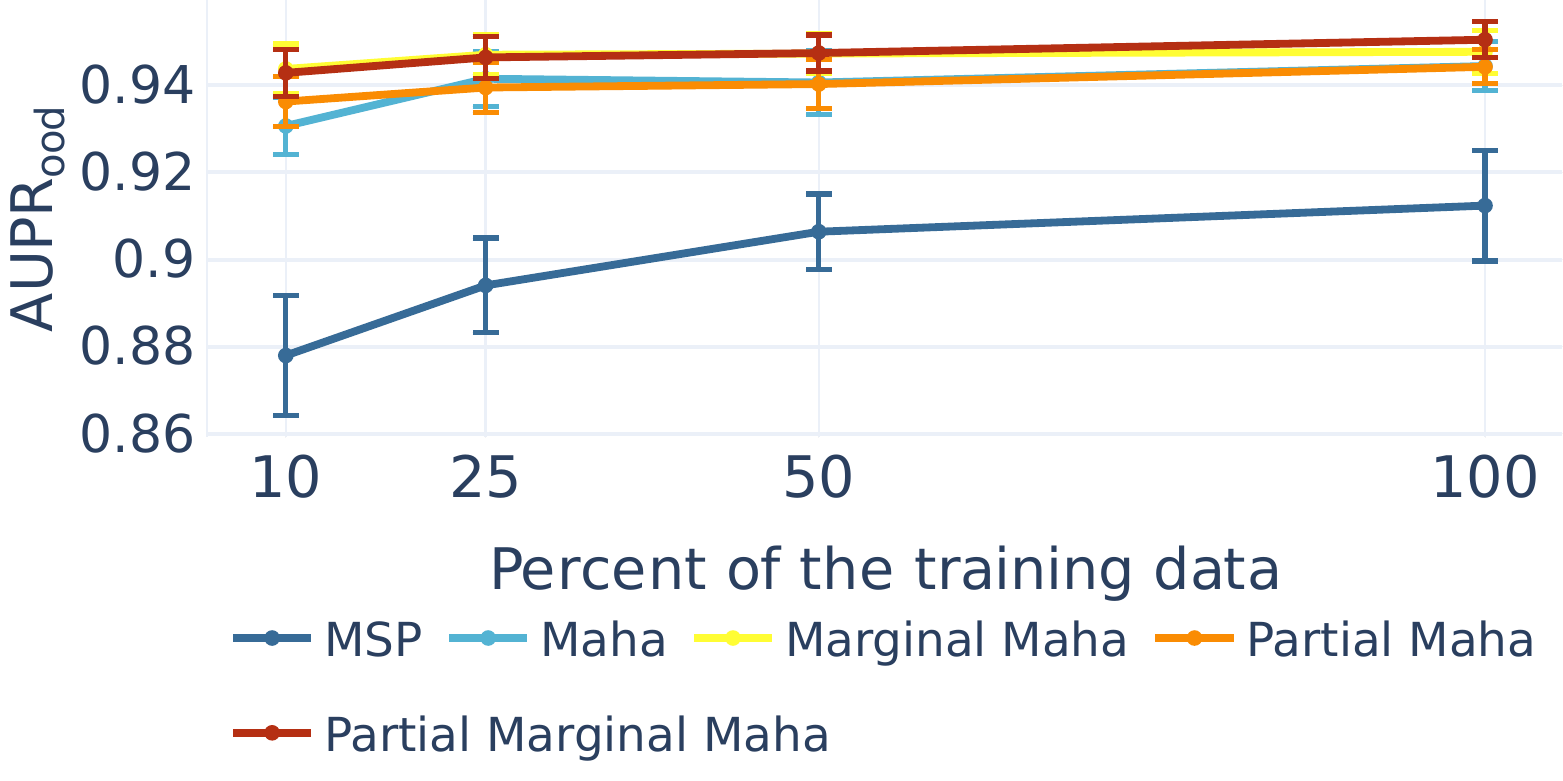}
    \caption{OX: fraction of train data used, CLINC150, OY: performance of OOD detection score. Mahalanobis distance and its variants need less data for OOD detection.}
    \label{fig:dataset_size_var}
\end{figure}

All Mahalanobis distance variants outperform Euclidean distance by far; see Fig.~\ref{fig:maha_var_comp}. Euclidean distance does not take the correlation between features into account. Although there is little difference between Mahalanobis distance variants, partial and marginal variants are more stable when varying training data size.  Marginal Mahalanobis distance is less affected by the reduction of training data; see \ref{fig:dataset_size_var}.

\section{Conclusion}

Out-of-Domain (OOD) detection task is becoming core to modern dialog systems. Successful detection and rejection of OOD utterances in real-life applications increase the dialog assistant's credibility and improves user experience. This paper compared multiple techniques for unsupervised OOD detection, applied to three commonly used NLU datasets, in particular, CLINC150, ROSTD, and SNIPS. We exploited different text representation models, ranging from the old-fashioned bag-of-word modes to the most recent pre-trained Transformers. We adopted best practices used in the vision domain and previously established state-of-the-art methods within the scope of unsupervised methods, namely, Maximum Softmax Probability, Likelihood Ratio, and Mahalanobis distance, along with its modifications.  

With the help of Transformer-based models, equipped with Mahalanobis distance, we establish new state-of-the-art results. To that end, we show that fine-tuning with ID data's supervision plays a crucial role, allowing re-shaping, favorable for the task, of the embedding space. These results are supported in line with \cite{reimers2019sentence}, confirming that fine-tuning Transformers improves the performance of the downstream unsupervised tasks. The proposed pipeline, i.e., fine-tuning a Transformer and using Mahalanobis distance, is robust to distillation. Supporting smaller models is essential for edge devices, where distilled models are usually deployed. Reduced in size, distilled versions of pre-trained Transformers models perform on par with the full-size models. Mahalanobis distance remains stable, even when used with a distilled model.

Still, there are some limitations to the Mahalanobis OOD score. In the first place, it depends on the geometrical features of the embedding space, which could be spoilt if, for example, the embedder is used simultaneously as a classification model and overfits. The greatest challenge is then semantically similar utterances, of which one is in ID, and the other is OOD. For example, this can happen if the dialog assistant supports only one of two related actions. Future research directions should consider such cases and the trade-off between the accuracy of intents classification and OOD detection.

\section*{Acknowledgments}
Ekaterina Artemova is partially supported by the framework of the HSE University Basic Research Program and funded by the Russian Academic Excellence Project  ``5-100''.

\bibliography{references}

\begin{thebibliography}{36}
\providecommand{\natexlab}[1]{#1}
\providecommand{\url}[1]{\texttt{#1}}
\providecommand{\urlprefix}{URL }
\expandafter\ifx\csname urlstyle\endcsname\relax
  \providecommand{\doi}[1]{doi:\discretionary{}{}{}#1}\else
  \providecommand{\doi}{doi:\discretionary{}{}{}\begingroup
  \urlstyle{rm}\Url}\fi

\bibitem[{Blundell et~al.(2015)Blundell, Cornebise, Kavukcuoglu, and
  Wierstra}]{blundell2015weight}
Blundell, C.; Cornebise, J.; Kavukcuoglu, K.; and Wierstra, D. 2015.
\newblock Weight Uncertainty in Neural Network.
\newblock In \emph{International Conference on Machine Learning}, 1613--1622.

\bibitem[{De~Maesschalck, Jouan-Rimbaud, and Massart(2000)}]{de2000mahalanobis}
De~Maesschalck, R.; Jouan-Rimbaud, D.; and Massart, D.~L. 2000.
\newblock The Mahalanobis distance.
\newblock \emph{Chemometrics and intelligent laboratory systems} 50(1): 1--18.

\bibitem[{Devlin et~al.(2019)Devlin, Chang, Lee, and
  Toutanova}]{devlin2019bert}
Devlin, J.; Chang, M.-W.; Lee, K.; and Toutanova, K. 2019.
\newblock BERT: Pre-training of Deep Bidirectional Transformers for Language
  Understanding.
\newblock In \emph{NAACL-HLT (1)}.

\bibitem[{Falcon(2019)}]{falcon2019pytorch}
Falcon, W. 2019.
\newblock PyTorch Lightning.
\newblock \emph{GitHub. Note:
  https://github.com/PyTorchLightning/pytorch-lightning Cited by} 3.

\bibitem[{{Gangal} et~al.(2019){Gangal}, {Arora}, {Einolghozati}, and
  {Gupta}}]{ll_ratio_nlp_facebook}
{Gangal}, V.; {Arora}, A.; {Einolghozati}, A.; and {Gupta}, S. 2019.
\newblock {Likelihood Ratios and Generative Classifiers for Unsupervised
  Out-of-Domain Detection In Task Oriented Dialog}.
\newblock \emph{arXiv e-prints} arXiv:1912.12800.

\bibitem[{Ghosal et~al.(2018)Ghosal, Sonam, Saha, Ekbal, and
  Bhattacharyya}]{ghosal2018investigating}
Ghosal, T.; Sonam, R.; Saha, S.; Ekbal, A.; and Bhattacharyya, P. 2018.
\newblock Investigating domain features for scope detection and classification
  of scientific articles.
\newblock In \emph{Proceedings of the Eleventh International Conference on
  Language Resources and Evaluation (LREC 2018)}, 7--12.

\bibitem[{Gu, Akoglu, and Rinaldo(2019)}]{Gu2019StatisticalAO}
Gu, X.; Akoglu, L.; and Rinaldo, A. 2019.
\newblock Statistical Analysis of Nearest Neighbor Methods for Anomaly
  Detection.
\newblock In \emph{NeurIPS}.

\bibitem[{Harris(1954)}]{haris_bow}
Harris, Z.~S. 1954.
\newblock Distributional structure.
\newblock \emph{Word} 10: 146--162.

\bibitem[{{Hendrycks} and {Gimpel}(2016)}]{msp}
{Hendrycks}, D.; and {Gimpel}, K. 2016.
\newblock {A Baseline for Detecting Misclassified and Out-of-Distribution
  Examples in Neural Networks}.
\newblock \emph{arXiv e-prints} arXiv:1610.02136.

\bibitem[{Hendrycks, Lee, and Mazeika(2019)}]{hendrycks_pretraining19}
Hendrycks, D.; Lee, K.; and Mazeika, M. 2019.
\newblock Using Pre-Training Can Improve Model Robustness and Uncertainty.
\newblock In Chaudhuri, K.; and Salakhutdinov, R., eds., \emph{Proceedings of
  the 36th International Conference on Machine Learning}, volume~97 of
  \emph{Proceedings of Machine Learning Research}, 2712--2721. Long Beach,
  California, USA: PMLR.
\newblock \urlprefix\url{http://proceedings.mlr.press/v97/hendrycks19a.html}.

\bibitem[{Hendrycks et~al.(2020)Hendrycks, Liu, Wallace, Dziedzic, Krishnan,
  and Song}]{Hendrycks2020PretrainedTI}
Hendrycks, D.; Liu, X.; Wallace, E.; Dziedzic, A.; Krishnan, R.; and Song,
  D.~X. 2020.
\newblock Pretrained Transformers Improve Out-of-Distribution Robustness.
\newblock In \emph{ACL}.

\bibitem[{Hendrycks, Mazeika, and Dietterich(2019)}]{hendrycks2018deep_OE}
Hendrycks, D.; Mazeika, M.; and Dietterich, T. 2019.
\newblock Deep Anomaly Detection with Outlier Exposure.
\newblock In \emph{International Conference on Learning Representations}.
\newblock \urlprefix\url{https://openreview.net/forum?id=HyxCxhRcY7}.

\bibitem[{Kamath, Jia, and Liang(2020)}]{kamath2020selective}
Kamath, A.; Jia, R.; and Liang, P. 2020.
\newblock Selective Question Answering under Domain Shift.
\newblock In \emph{Proceedings of the 58th Annual Meeting of the Association
  for Computational Linguistics}, 5684--5696.

\bibitem[{Kamoi and Kobayashi(2020)}]{Kamoi2020WhyIT}
Kamoi, R.; and Kobayashi, K. 2020.
\newblock Why is the Mahalanobis Distance Effective for Anomaly Detection?
\newblock \emph{ArXiv} abs/2003.00402.

\bibitem[{Larson et~al.(2019{\natexlab{a}})Larson, Mahendran, Peper, Clarke,
  Lee, Hill, Kummerfeld, Leach, Laurenzano, Tang, and Mars}]{clinc150}
Larson, S.; Mahendran, A.; Peper, J.~J.; Clarke, C.; Lee, A.; Hill, P.;
  Kummerfeld, J.~K.; Leach, K.; Laurenzano, M.~A.; Tang, L.; and Mars, J.
  2019{\natexlab{a}}.
\newblock An Evaluation Dataset for Intent Classification and Out-of-Scope
  Prediction.
\newblock In \emph{Proceedings of the 2019 Conference on Empirical Methods in
  Natural Language Processing and the 9th International Joint Conference on
  Natural Language Processing (EMNLP-IJCNLP)}.
\newblock \urlprefix\url{https://www.aclweb.org/anthology/D19-1131}.

\bibitem[{Larson et~al.(2019{\natexlab{b}})Larson, Mahendran, Peper, Clarke,
  Lee, Hill, Kummerfeld, Leach, Laurenzano, Tang et~al.}]{larson2019evaluation}
Larson, S.; Mahendran, A.; Peper, J.~J.; Clarke, C.; Lee, A.; Hill, P.;
  Kummerfeld, J.~K.; Leach, K.; Laurenzano, M.~A.; Tang, L.; et~al.
  2019{\natexlab{b}}.
\newblock An Evaluation Dataset for Intent Classification and Out-of-Scope
  Prediction.
\newblock In \emph{Proceedings of the 2019 Conference on Empirical Methods in
  Natural Language Processing and the 9th International Joint Conference on
  Natural Language Processing (EMNLP-IJCNLP)}, 1311--1316.

\bibitem[{Lee et~al.(2018)Lee, Lee, Lee, and Shin}]{Lee2018ASU}
Lee, K.; Lee, K.; Lee, H.; and Shin, J. 2018.
\newblock A Simple Unified Framework for Detecting Out-of-Distribution Samples
  and Adversarial Attacks.
\newblock In \emph{NeurIPS}.

\bibitem[{{Liang}, {Li}, and {Srikant}(2017)}]{odin}
{Liang}, S.; {Li}, Y.; and {Srikant}, R. 2017.
\newblock {Enhancing The Reliability of Out-of-distribution Image Detection in
  Neural Networks}.
\newblock \emph{arXiv e-prints} arXiv:1706.02690.

\bibitem[{{Lin} and {Xu}(2019)}]{intent_detection_margin_loss}
{Lin}, T.-E.; and {Xu}, H. 2019.
\newblock {Deep Unknown Intent Detection with Margin Loss}.
\newblock \emph{arXiv e-prints} arXiv:1906.00434.

\bibitem[{Liu et~al.(2020)Liu, Lin, Padhy, Tran, Bedrax-Weiss, and
  Lakshminarayanan}]{liu2020simple}
Liu, J.~Z.; Lin, Z.; Padhy, S.; Tran, D.; Bedrax-Weiss, T.; and
  Lakshminarayanan, B. 2020.
\newblock Simple and Principled Uncertainty Estimation with Deterministic Deep
  Learning via Distance Awareness.

\bibitem[{Liu et~al.(2019)Liu, Ott, Goyal, Du, Joshi, Chen, Levy, Lewis,
  Zettlemoyer, and Stoyanov}]{liu2019roberta}
Liu, Y.; Ott, M.; Goyal, N.; Du, J.; Joshi, M.; Chen, D.; Levy, O.; Lewis, M.;
  Zettlemoyer, L.; and Stoyanov, V. 2019.
\newblock Roberta: A robustly optimized bert pretraining approach.
\newblock \emph{arXiv preprint arXiv:1907.11692} .

\bibitem[{Malinin and Gales(2018)}]{NIPS2018_7936}
Malinin, A.; and Gales, M. 2018.
\newblock Predictive Uncertainty Estimation via Prior Networks.
\newblock In Bengio, S.; Wallach, H.; Larochelle, H.; Grauman, K.;
  Cesa-Bianchi, N.; and Garnett, R., eds., \emph{Advances in Neural Information
  Processing Systems 31}, 7047--7058. Curran Associates, Inc.
\newblock
  \urlprefix\url{http://papers.nips.cc/paper/7936-predictive-uncertainty-estimation-via-prior-networks.pdf}.

\bibitem[{Mandelbaum and Weinshall(2017)}]{Mandelbaum2017DistancebasedCS}
Mandelbaum, A.; and Weinshall, D. 2017.
\newblock Distance-based Confidence Score for Neural Network Classifiers.
\newblock \emph{ArXiv} abs/1709.09844.

\bibitem[{Murphy(2012)}]{murphy2012machine}
Murphy, K.~P. 2012.
\newblock Machine Learning: A Probabilistic Perspective.
\newblock \emph{The MIT Press} ISBN 0262018020.

\bibitem[{Nalisnick et~al.(2019)Nalisnick, Matsukawa, Teh, Gorur, and
  Lakshminarayanan}]{nalisnick2018do}
Nalisnick, E.; Matsukawa, A.; Teh, Y.~W.; Gorur, D.; and Lakshminarayanan, B.
  2019.
\newblock Do Deep Generative Models Know What They Don't Know?
\newblock In \emph{International Conference on Learning Representations}.
\newblock \urlprefix\url{https://openreview.net/forum?id=H1xwNhCcYm}.

\bibitem[{Orhan(2019)}]{orhan2019robustness}
Orhan, A.~E. 2019.
\newblock Robustness properties of Facebook's ResNeXt WSL models.

\bibitem[{Paszke et~al.(2019)Paszke, Gross, Massa, Lerer, Bradbury, Chanan,
  Killeen, Lin, Gimelshein, Antiga, Desmaison, Köpf, Yang, DeVito, Raison,
  Tejani, Chilamkurthy, Steiner, Fang, Bai, and Chintala}]{paszke2019pytorch}
Paszke, A.; Gross, S.; Massa, F.; Lerer, A.; Bradbury, J.; Chanan, G.; Killeen,
  T.; Lin, Z.; Gimelshein, N.; Antiga, L.; Desmaison, A.; Köpf, A.; Yang, E.;
  DeVito, Z.; Raison, M.; Tejani, A.; Chilamkurthy, S.; Steiner, B.; Fang, L.;
  Bai, J.; and Chintala, S. 2019.
\newblock PyTorch: An Imperative Style, High-Performance Deep Learning Library.

\bibitem[{Pennington, Socher, and Manning(2014)}]{pennington2014glove}
Pennington, J.; Socher, R.; and Manning, C.~D. 2014.
\newblock Glove: Global vectors for word representation.
\newblock In \emph{Proceedings of the 2014 conference on empirical methods in
  natural language processing (EMNLP)}, 1532--1543.

\bibitem[{Reimers and Gurevych(2019)}]{reimers2019sentence}
Reimers, N.; and Gurevych, I. 2019.
\newblock Sentence-BERT: Sentence Embeddings using Siamese BERT-Networks.
\newblock In \emph{Proceedings of the 2019 Conference on Empirical Methods in
  Natural Language Processing and the 9th International Joint Conference on
  Natural Language Processing (EMNLP-IJCNLP)}, 3973--3983.

\bibitem[{{Ren} et~al.(2019){Ren}, {Liu}, {Fertig}, {Snoek}, {Poplin},
  {DePristo}, {Dillon}, and {Lakshminarayanan}}]{ll_ratio_google}
{Ren}, J.; {Liu}, P.~J.; {Fertig}, E.; {Snoek}, J.; {Poplin}, R.; {DePristo},
  M.~A.; {Dillon}, J.~V.; and {Lakshminarayanan}, B. 2019.
\newblock {Likelihood Ratios for Out-of-Distribution Detection}.
\newblock \emph{arXiv e-prints} arXiv:1906.02845.

\bibitem[{Sanh et~al.(2019)Sanh, Debut, Chaumond, and
  Wolf}]{sanh2019distilbert}
Sanh, V.; Debut, L.; Chaumond, J.; and Wolf, T. 2019.
\newblock DistilBERT, a distilled version of BERT: smaller, faster, cheaper and
  lighter.

\bibitem[{{Schuster} et~al.(2018){Schuster}, {Gupta}, {Shah}, and
  {Lewis}}]{rostd_indomain}
{Schuster}, S.; {Gupta}, S.; {Shah}, R.; and {Lewis}, M. 2018.
\newblock {Cross-Lingual Transfer Learning for Multilingual Task Oriented
  Dialog}.
\newblock \emph{arXiv e-prints} arXiv:1810.13327.

\bibitem[{Tan et~al.(2019)Tan, Yu, Wang, Wang, Potdar, Chang, and
  Yu}]{tan2019out}
Tan, M.; Yu, Y.; Wang, H.; Wang, D.; Potdar, S.; Chang, S.; and Yu, M. 2019.
\newblock Out-of-Domain Detection for Low-Resource Text Classification Tasks.
\newblock In \emph{Proceedings of the 2019 Conference on Empirical Methods in
  Natural Language Processing and the 9th International Joint Conference on
  Natural Language Processing (EMNLP-IJCNLP)}, 3557--3563.

\bibitem[{Wolf et~al.(2019)Wolf, Debut, Sanh, Chaumond, Delangue, Moi, Cistac,
  Rault, Louf, Funtowicz et~al.}]{wolf2019huggingface}
Wolf, T.; Debut, L.; Sanh, V.; Chaumond, J.; Delangue, C.; Moi, A.; Cistac, P.;
  Rault, T.; Louf, R.; Funtowicz, M.; et~al. 2019.
\newblock HuggingFace's Transformers: State-of-the-art Natural Language
  Processing.
\newblock \emph{ArXiv} arXiv--1910.

\bibitem[{Yilmaz and Toraman(2020)}]{yilmaz2020kloos}
Yilmaz, E.~H.; and Toraman, C. 2020.
\newblock KLOOS: KL Divergence-based Out-of-Scope Intent Detection in
  Human-to-Machine Conversations.
\newblock In \emph{Proceedings of the 43rd International ACM SIGIR Conference
  on Research and Development in Information Retrieval}, 2105--2108.

\bibitem[{{Zheng}, {Chen}, and {Huang}(2020)}]{zheng2020out}
{Zheng}, Y.; {Chen}, G.; and {Huang}, M. 2020.
\newblock Out-of-domain detection for natural language understanding in dialog
  systems.
\newblock In \emph{IEEE/ACM Transactions on Audio, Speech, and Language
  Processing}, volume~28, 1198--1209. IEEE.

\end{thebibliography}

\end{document}